\documentclass[twocolumn,letterpaper]{IEEEAerospaceCLS}  

\usepackage[]{graphicx}    
\usepackage{amsfonts}
\usepackage{amsmath}
\newcommand{\ignore}[1]{}  
\newcommand{\vect}[1]{\boldsymbol{#1}}

\usepackage{xcolor}

\graphicspath{{images/}}

\begin{document}
\title{Geometric Cross-Modal Comparison \\ of Heterogeneous Sensor Data}

\author{%
Christopher J. Tralie\\
Department of Electrical \\
and Computer Engineering\\b
Duke University
\and
Abraham Smith\\
Geometric Data Analytics, Inc.\\
Department of Mathematics\\
University of Wisconsin-Stout
\and
Nathan Borggren\\
Geometric Data Analytics, Inc.
\and
Jay Hineman\\
Geometric Data Analytics, Inc.
\and
Paul Bendich\\
Geometric Data Analytics, Inc.\\
Department of Mathematics \\
Duke University
\and
Peter Zulch \\
Air Force Research Laboratory 
\and
John Harer\\
Geometric Data Analytics, Inc.\\
Department of Mathematics\\
Duke University
}

\maketitle

\thispagestyle{plain}
\pagestyle{plain}

\begin{abstract}
In this work, we address the problem of cross-modal comparison of aerial data streams.
A variety of simulated automobile trajectories are sensed using two different modalities: full-motion video, and radio-frequency (RF) signals received by detectors at various locations.
The information represented by the two modalities is compared using self-similarity matrices (SSMs) corresponding to time-ordered point clouds in feature spaces of each of these data sources; we note that these feature spaces can be of entirely different scale and dimensionality.
Several metrics for comparing SSMs are explored, including a cutting-edge time-warping technique that can simultaneously handle local time warping and partial matches, while also controlling for the change in geometry between feature spaces of the two modalities.
We note that this technique is quite general, and does not depend on the choice of modalities.
In this particular setting, we demonstrate that the cross-modal distance between SSMs corresponding to the same trajectory type is smaller than the cross-modal distance between SSMs corresponding to distinct trajectory types, and we formalize this observation via precision-recall metrics in experiments.
Finally, we comment on promising implications of these ideas for future integration into multiple-hypothesis tracking systems.
\end{abstract}

\tableofcontents

\section{Introduction}

We advocate the use of a geometric construct called \emph{Self-Similarity Matrices} (SSMs, Section \ref{sec:SSM}) in the comparison of cross-modal data
sensed from disparate aerial data streams. Unlike the results of deep network based approaches (e.g., \cite{Sarfraz2016}), the reasons for the cross-modal similarities found
by the SSMs are immediately apparent to the user.  
This paper presents a series of proof-of-principle experiments (Section \ref{sec:ExpRes}) that compare full-motion video (FMV) to combinations of RF signals received by receivers at many different locations, but we expect that this technique, which does not depend on the choice of specific modalities to compare, will prove quite general in the future.

The SSM is a two-dimensional matrix summary of a time-series of data.
This data can certainly be a single signal, such as the speed function associated to a snippet of vehicle track. But much more generally, it can be time-series data of any dimension, such as an ordered sequence of time-series, each representing RF information collected from one of several sensors.
The  $(i,j)$ entry of the SSM is the similarity between the data collected at times $t_i$ and $t_j$, computed using a similarity measure appropriate for that modality.
Different SSMs, \emph{including those arising from completely different modalities}, can then be compared using these matrices.
A key feature of general SSMs is that they are \emph{isometry-blind}: that is, they are completely invariant to rigid motions such as reflections and rotations, a fact which allows the analyst to completely ignore otherwise difficult alignment problems.

Several experiments with synthetic data (Section \ref{sec:Data}) are performed in this paper.
The main idea is that we simulate a large number of automobile trajectories corresponding to certain typical motions: as seen in Figure \ref{fig:Trajectories}, we focus on ``going straight at constant speed'' and
``switching lanes and then exiting the highway'' and ``making a u-turn.''
These scenes are observed using simulations of FMV, with added noise, to measure magnitude-of-velocity in a fixed coordinate systems.
The vehicles are also programmed to emit radio signals, and Doppler-shifted versions of these signals are received, again with noise,
by RF sensors positioned at random fixed locations throughout the scene.

The potential benefits for using SSMs in analyzing heterogenous sensor data can be seen in (for example) the top row of Figures \ref{fig:DopplerSimExample1} and \ref{fig:DopplerSimExample2}, where we observe clear qualitative similarities between SSMs from different modalities that correspond to the same scene, and clear qualitative differences for distinct scenes.
Much of this paper involves an exploration (Section \ref{sec:CSSM}) of several potential metrics for constructing, normalizing, and comparing SSMs, including a cutting-edge technique recently developed \cite{tralie2017warping} by the first author. The main conclusion is that these metrics can, with high but varying degrees of success, classify typical motions using only SSMs, and that the new technique does perform the best by a slight margin; see Figures \ref{fig:PRIBDTWs} and \ref{fig:PRPDs}.
Another key outcome can be seen by comparing the top rows of Figure \ref{fig:DopplerSimExample1} and \ref{fig:DopplerSimExample2}.
These show that SSMs in higher-dimensional spaces (taking all RF sensors into account at once) are much more stable to location error than SSMs corresponding to individual sensors.
As a consequence, we conjecture that SSMs from higher-dimensional spaces will prove useful as a summary feature for the \emph{situation assessment} (e.g. \cite{blasch2012}) step within 
the fusion of data from a large number of heterogenous sensor sources, as well as from a large number of sensors sensing the same modality from different locations within a scene.

Finally, we observe that there is wide awareness \cite{chavali2014} within the tracking community that incorporating sensed data from different modalities can improve tracker performance.
To cite just one practical example, Nayak et al \cite{Nayak2008} construct an algorithm which uses coarse tracking information from a multitude of cheap audio sensors and uses it to guide the deployment
of a smaller set of more expensive video cameras.
Although this paper does not pursue this beyond the notional stage, Section \ref{sec:MHT} outlines a future integration of the ideas presented here into multi-target tracking systems.

\section*{Acknowledgments}
This paper under contract STTR \# FA8750-16-C-0220 has been approved for public distribution by the Air Force, case \#88ABW-2017-5856;
Bendich, Borggren, Hineman, Smith, and Harer were partially supported under this same contract by the Air Force Research Laboratory.
Bendich, Harer, and Tralie were also partially supported by the National Science Foundation BIGDATA grant, \# DMS 144749.
Tralie was also partially supported by an NSF Graduate Fellowship NSF under grant \# DGF-1106401.
Justin Curry is thanked for helpful discussions about self-similarity matrices.

\section{Self-Similarity Matrices}
\label{sec:SSM}


\begin{figure}
\centering
\includegraphics[width=\columnwidth]{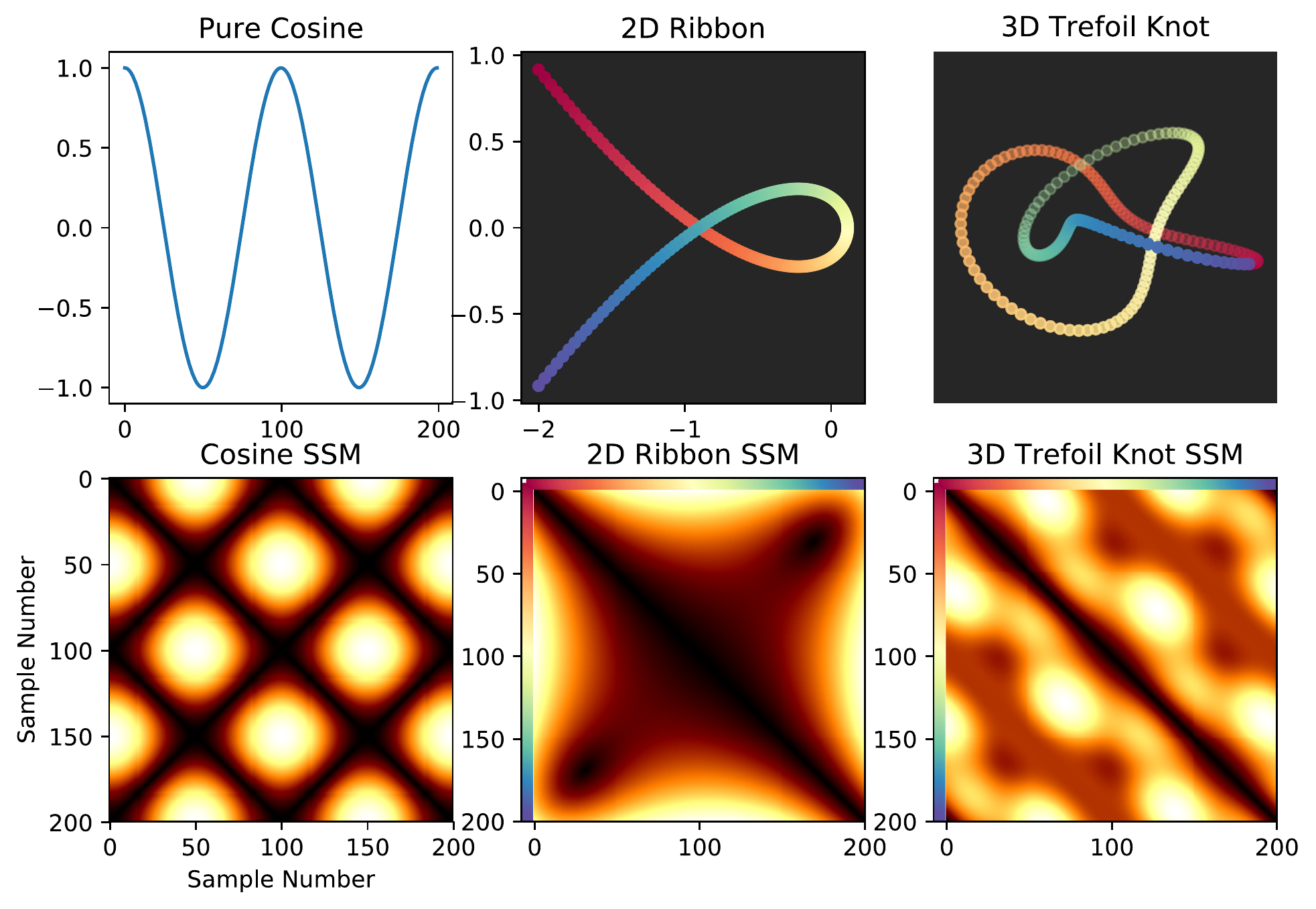}
\caption{An example of self-similarity matrices (SSMs) built on 3 different time-ordered point clouds (TOPCs): A 1D cosine, a 2D ribbon, and a 3D knot.  For the 2D and 3D point clouds, the color on the point cloud plot corresponds to the color label on the sample axes of the SSMs. }
\label{fig:SSMConcept}
\end{figure}


Given a space curve $\gamma: [0, 1] \rightarrow R^d$ defining a trajectory of arbitrary type in a Euclidean space of arbitrary dimension $d$,  a {\em self-similarity image} (SSI) is a function $D: [0, 1] \times [0, 1] \rightarrow R$ so that $D_{\gamma}(i, j) = d(\gamma(i), \gamma(j))$.  In other words, a self-similarity image contains the distance between every pair of points on the curve, as measured in the Euclidean vector space in which the curve lives.  In practice, we deal with discretized curves, which we refer to as ``time-ordered point clouds'' (TOPCs); that is, an N-length, $d$-dimensional time-ordered point cloud is a sequence of vectors in $R^d$, $X_1, X_2, \hdots, X_N$, indexed by time.  The discrete version of SSIs for time-ordered point clouds is the {\em Self-Similarity Matrix} (SSM) $D$, so that $D_{ij} = ||X_i - X_j||_2$, and $D$ is a symmetric $N \times N$ matrix for a point cloud with $N$ points.  Figure~\ref{fig:SSMConcept} shows an example of 3 different sampled curves, each defined in a different dimension, and their corresponding SSMs.  Note that for the 1D point cloud, the Euclidean metric between two different samples is simply the absolute difference between their heights.

One of the most attractive properties of SSMs is that they are {\em invariant to isometries}.  In the case of SSMs on 1D time-ordered point clouds, this means they remain the same regardless of whether the signal has been reflected from top to bottom or translated up and down.  For higher dimensional point clouds, this means that SSMs remain the same under any rotation, translation, flip, or combination therein.  Since time-evolving processes can often be modeled in feature spaces, and since one often only cares about {\em relative} changes in that feature space, SSMs are an invaluable tool across many domains.  For instance, the authors in \cite{junejo2008cross} compute SSMs on video feature spaces, which they use to recognize actions from different points of view, which are approximate isometries in the feature spaces.  Other works have used SSMs to model changes in music, since structural progressions can be thought of as relative shape changes \cite{foote2000automatic,bello2009SSMStructure,mcfee2014analyzing}, or, as in the video activity case, since musical expressions have approximately the same SSM when performed with different instruments \cite{tralie2015cover}.  In the dynamical systems community, SSMs are used to compute recurrence statistics to detect periodicity, chaos, and entropy \cite{mcguire1997recurrence}.

\begin{figure}
\centering
\includegraphics[width=\columnwidth]{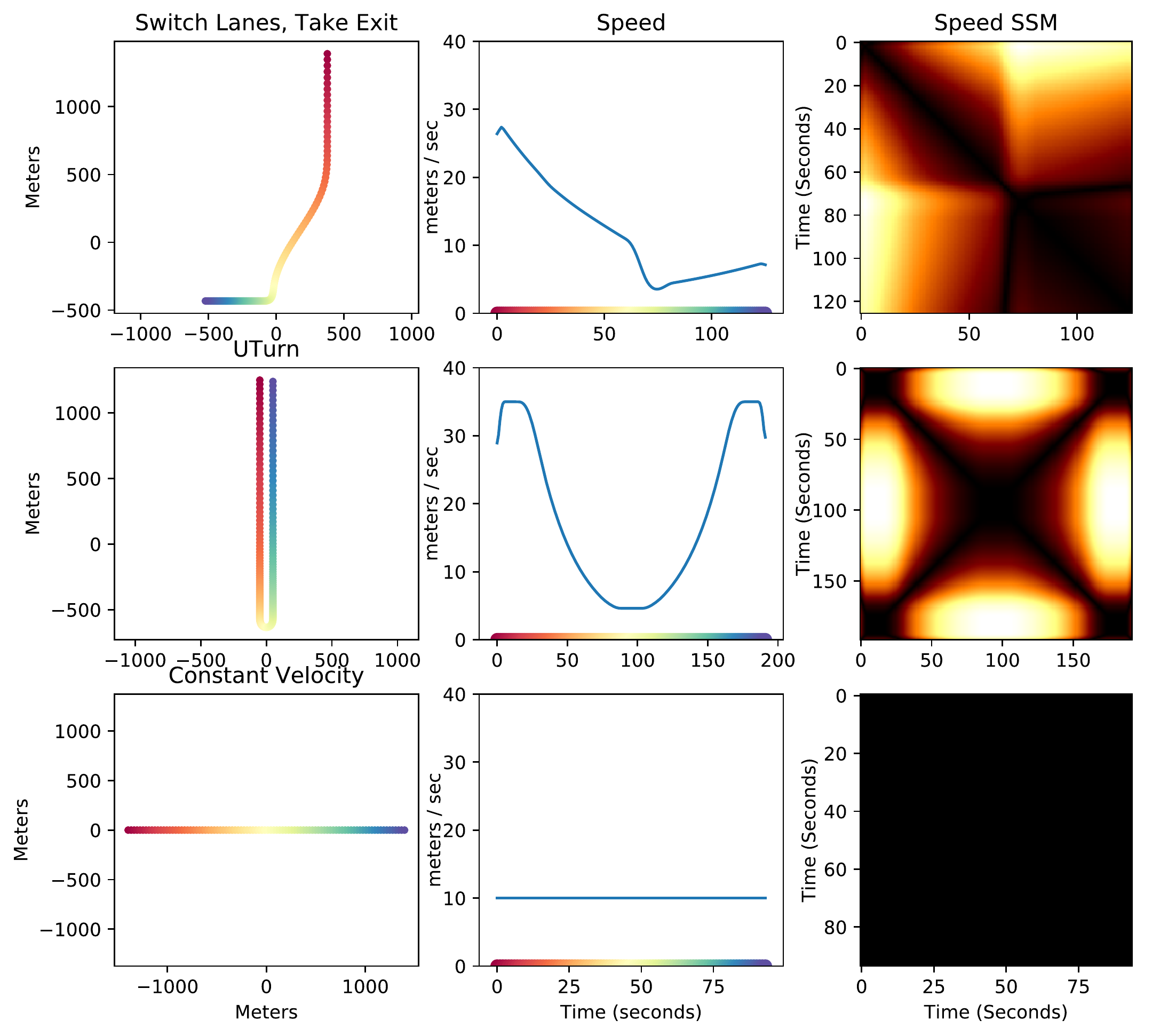}
\caption{The three typical motions we consider in our simulations.
Each row is a motion. The first column shows the plotted trajectory with time moving from red to blue, the second shows speed against time, and the third displays the speed SSM.
}
\label{fig:Trajectories}
\end{figure}

In this work, we use SSMs as a proxy for comparing RF signals to estimated speed profiles.  Figure~\ref{fig:Trajectories} shows the three example trajectories we use throughout this paper, along with the SSMs corresponding to their 1D speed profiles.  We will show that although RF signals are a different modality, their SSMs are similar in character to the speed SSMs. These observatons will be rigorously quantified via the metrics discussed in Section \ref{sec:CSSM}, leading to the supervised-learning experiments in Section \ref{sec:ExpRes}.

\section{Data and Sensing Modalities}
\label{sec:Data}

\begin{figure*}
\centering
\includegraphics[width=\textwidth]{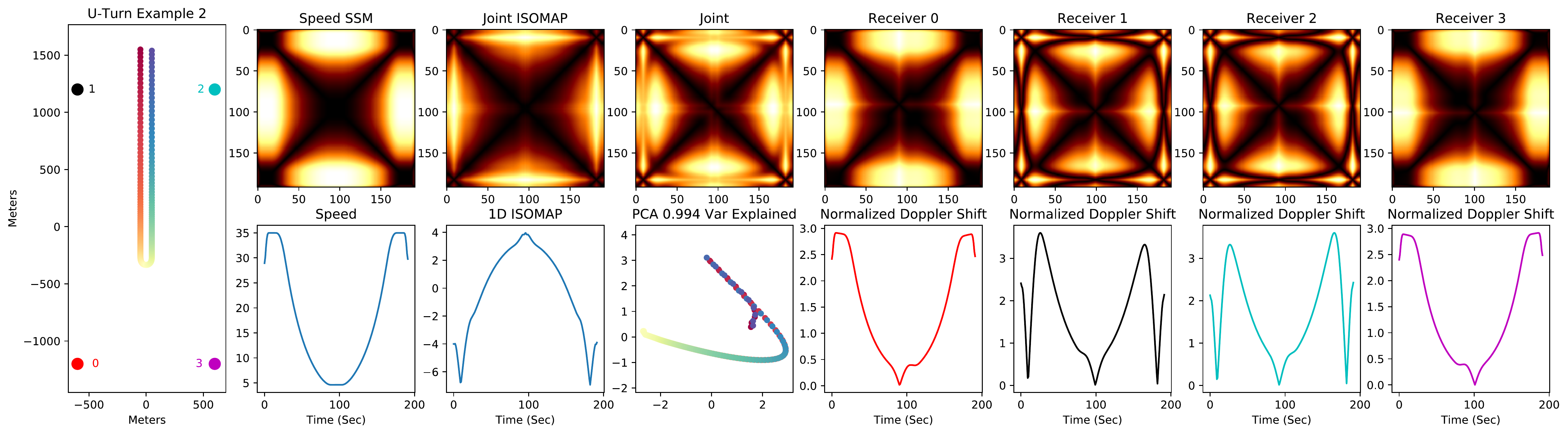}
\caption{An example of the speed profile and Doppler shifts from 4 different receivers, along with SSMs of 4D joint embeddings and ISOMAP of those embeddings.  The trajectory is drawn on the left, and the receiver positions are drawn and numbered as colored dots.  The SSMs are all on the top row, and the normalized 1D Doppler shifts from the 4 receivers are drawn on the bottom, in addition to the speed profile (second plot on the bottom left), 1D ISOMAP (third plot), and 2D PCA of the joint embedding (fourth plot)}
\label{fig:DopplerSimExample1}
\end{figure*}

\begin{figure*}
\centering
\includegraphics[width=\textwidth]{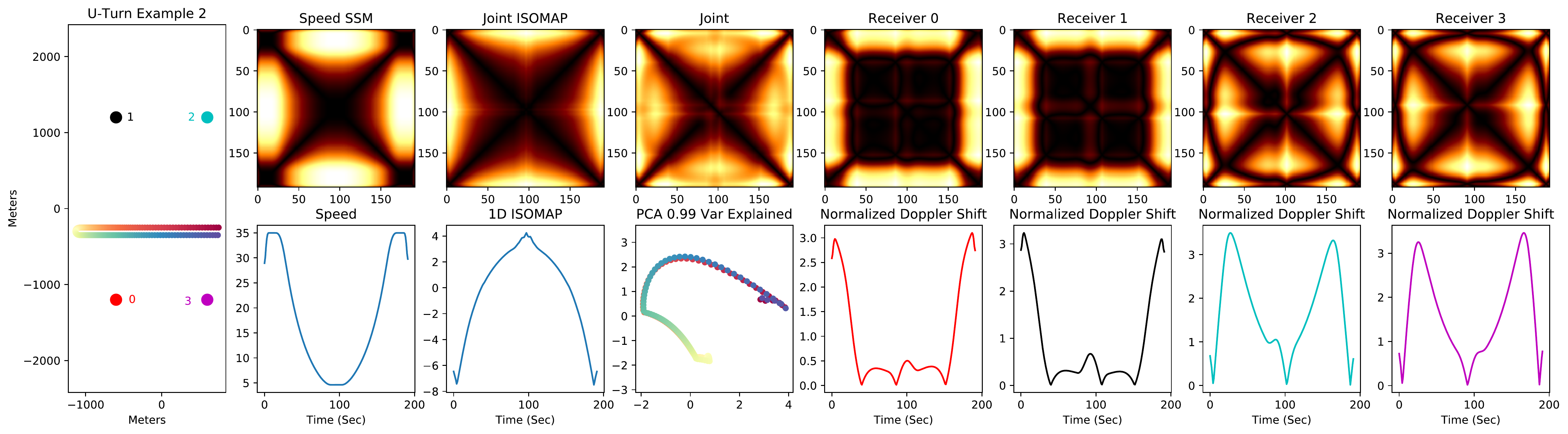}
\caption{Same as Figure~\ref{fig:DopplerSimExample1}, except the transmitter has been moved to a different location.  Note that individual received Doppler can change drastically, but the joint embedding and ISOMAP are more stable.}
\label{fig:DopplerSimExample2}
\end{figure*}

\begin{figure*}
\centering
\includegraphics[width=\textwidth]{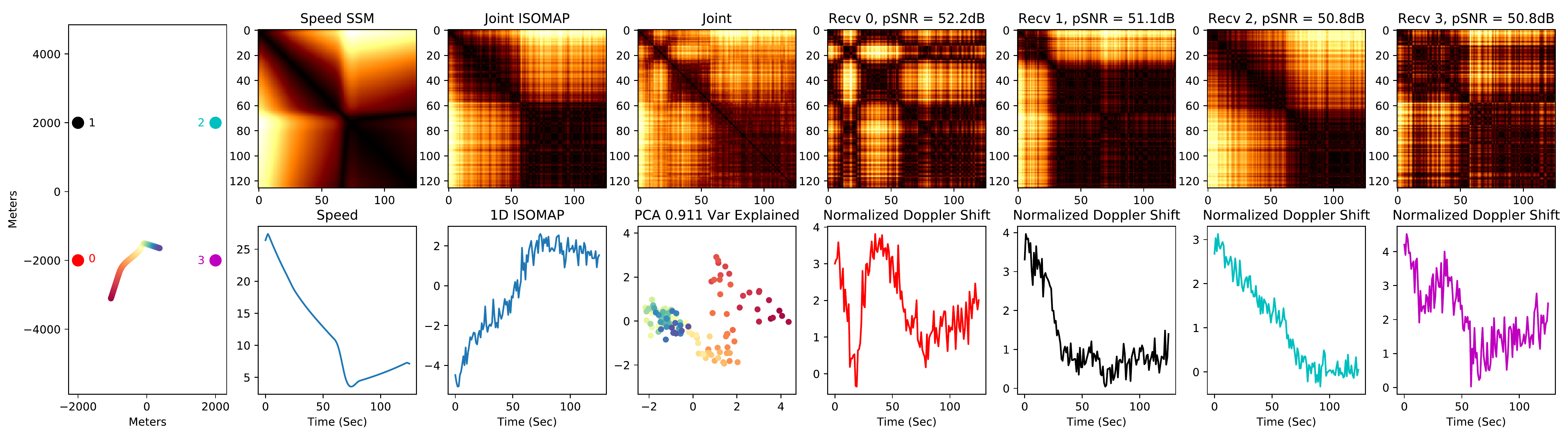}
\caption{An example of the Doppler SSMs for ``taking an exit'' at an average pSNR of about 51dB over all of the receivers.  The trajectory is arbitrarily rotated and translated with respect to 4 sensors.}
\label{fig:NoiseExample}
\end{figure*}


To showcase our multimodal comparison schemes, we run a very simple simulation of Doppler receivers acting on the trajectories shown in the left column of  Figure~\ref{fig:Trajectories}.  The radio signals are received by multiple detectors, and the signal at each detector is computed using the usual non-relativistic Doppler equation
\[ f_{\text{recvd}}(t_{i'}) = \left(1 - \frac{v(t)}{c}\right)
f_{\text{trans}}(t_i)\]
where $r(t_i)$ is the distance from the detector to the vehicle, and $\dot{r}(t_i) =
v(t_i)$ is the radial speed of the transmitter at time $t_i$.  Note that the
received signal experiences a slight delay, $t_{i'} = t_{i} + r(t_i)/d$.

The SSM of the received signal at each receiver is computed as
\[ S_{i,j} = | f_{\text{recvd}}(t_{i'}) - f_{\text{recvd}}(t_{j'})|,\qquad
\forall i,j\]

just as the pure cosine example in Figure~\ref{fig:SSMConcept}, though we will discuss higher dimensional SSMs resulting from joint embeddings in Section~\ref{sec:CSSM} and show benefits of such an embedding experimentally in Section~\ref{sec:ExpRes}.  Note that the computation of the SSM does \emph{not} require knowledge of the
distance, speed, or frequency of the transmitting vehicle.   Therefore, simulations allow us to build a general dictionary to reconstruct possible vehicle trajectories from the observed data; we comment on the future benefits of this capability in Section \ref{sec:MHT}.

\subsection{Noise Model}

We add Gaussian noise to our receivers, which we quantify as follows.  Define the peak signal to noise ratio (PSNR) as

\[PSNR = 10 \log_{10}(MAX^2/MSE) \]

where we define MAX as the maximum Doppler shift expected from a vehicle heading straight towards a receiver at 50 meters / second ($\approx 112$ Mph), and the mean squared error (MSE) is with respect to the Doppler shift that the corresponding sensor would receive from an unperturbed trajectory.  Figure~\ref{fig:NoiseExample} shows an example draw from the ``Taking Exit'' behavior, which includes a translation/rotation/flip and added noise.  Note that the noise added to the Doppler shift comes in addition to the radial perturbations in the sensors that naturally occur due to differing relative positions between the receivers and the trajectory, so there are multiple sources of confusion when comparing ideal speed profiles to Doppler signals.

\section{Constructing And Comparing SSMs}
\label{sec:CSSM}

This section discusses the different aspects of a computational pipeline that begins with aerial data streams and ends with quantifiable comparisons between SSMs.
The steps in the pipeline are: construct the SSM from the data streams, normalize the SSMs in such a way that intelligent cross-modal comparisons can be made, and define and then compute metrics between pairs of SSMs.
For each step, there are several possibilities, all of which are discussed in this subsection and which are then evaluated in the experiments in Section \ref{sec:ExpRes}.

\subsection{SSM Construction}
\label{sec:construction}

In the case of matching speed profiles to Doppler shifts from multiple sensors, there are a few options for how to generate SSMs from the raw data.  First, we need to normalize each of the Doppler traces to be on the same scale so that we can combine them equally without any one dominating.  We do this by simply dividing each by its standard deviation before processing.  The right four columns of Figure~\ref{fig:DopplerSimExample1} and Figure~\ref{fig:DopplerSimExample2} show the normalized Doppler shifts and their corresponding SSMs.  After this, a naive option to fuse them is to compute SSMs for each individual Doppler shift time series, and then to average them together.  A slightly more sophisticated approach is to consider a joint embedding of the time series from each of $d$ sensors; that is, each Doppler shift signal is mapped to a single coordinate in a $d$ dimensional space.  We can then compute an SSM in this space.  Note that since SSMs are constructed with pairwise distances only, it does not matter which dimension contains which receiver.  In other words, the pairwise distance is invariant to permutations of the orders in which the receivers are considered.

Column 4 of Figures~\ref{fig:DopplerSimExample1} and \ref{fig:DopplerSimExample2} show the SSM and a projection of the point cloud in 4 dimensional space .  In other words, resulting from a joint embedding of Doppler profiles from 4 sensors.  Note how although individual Doppler shifts change between Figure~\ref{fig:DopplerSimExample1} and Figure~\ref{fig:DopplerSimExample2}, the joint embedding is more stable, which is one of the primary benefits of this approach.

\begin{figure}
\centering
\includegraphics[width=0.6\columnwidth]{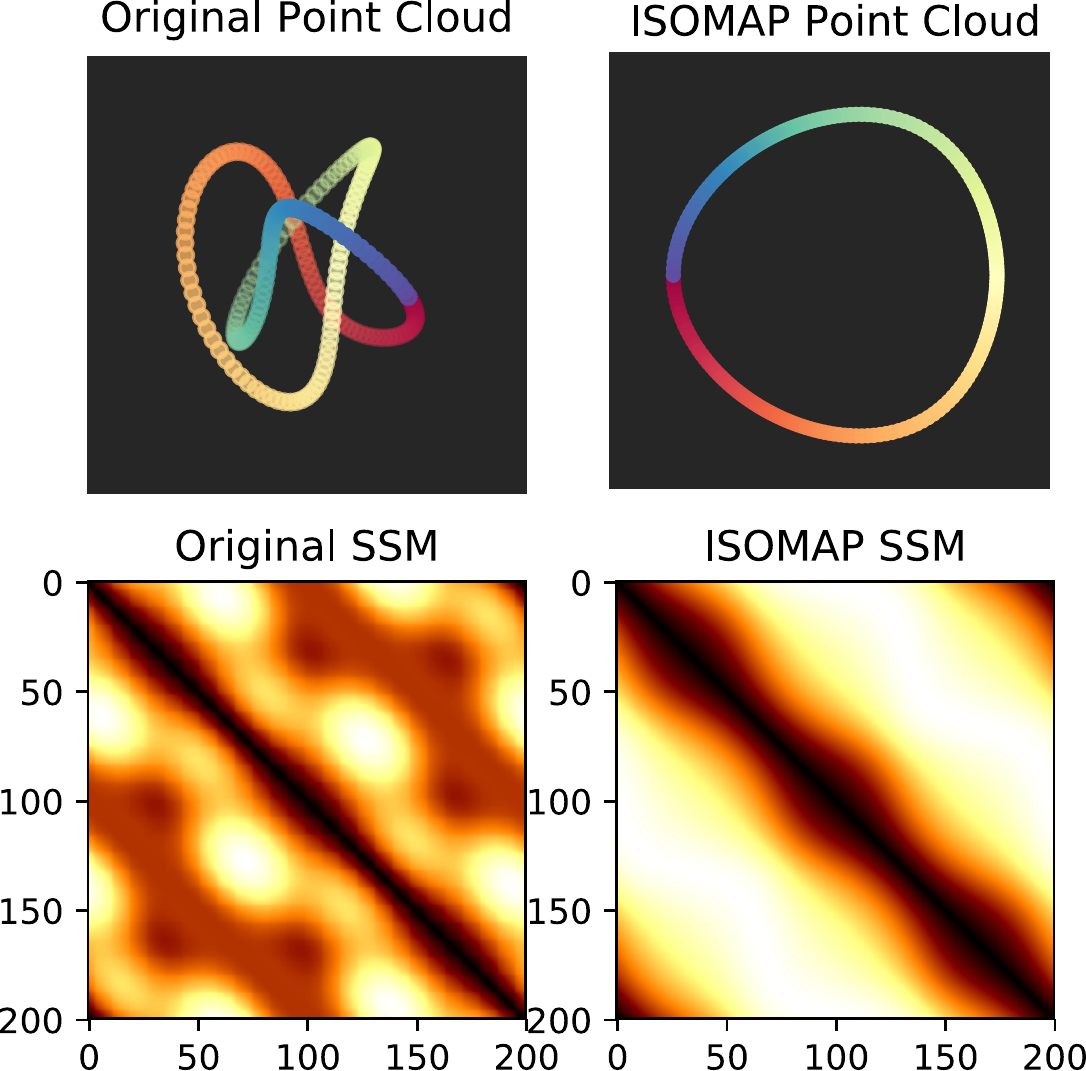}
\caption{An example of running 2D ISOMAP on a trefoil knot.  The algorithm successfully discovers the intrinsic loop structure of the knot.}
\label{fig:ISOMAPUnknotting}
\end{figure}

If it were the case that for every speed profile $s(t)$, that every Doppler signal $x(t)$ was a simple affine transformation $x = as + b$, for some constants $a$ and $b$, then a joint embedding of $d$ Doppler sensors would be

\[\vect{X}(t) = (a_1s(t) + b_1, a_2s(t) + b_2, ..., a_2s(t) + b_d)\]

which is $\vect{X}(t) = \vect{a}t + \vect{b}$ for some constant vectors $\vect{a}, \vect{b} \in R^d$, which is simply a line segment.  In practice, due to the fact that Doppler sensors pick up on changes in radial velocity instead of velocity, not every sensor is necessarily an affine scale of the speed or of any of the other sensors.  This means that the embedded geometry is usually distorted and not perfectly straight.  Principal component analysis \cite{Jolliffe1986}, a linear dimension reduction technique (row 2, column 4 of Figure~\ref{fig:DopplerSimExample1} and Figure~\ref{fig:DopplerSimExample2}), reveals that the geometry does indeed deviate from a straight line.  However, it is possible to factor out this local curvature using a a nonlinear dimension reduction technique known as ISOMAP \cite{tenenbaum2000global}.  ISOMAP works by building a nearest neighbor graph on the point cloud and replacing Euclidean distances with graph geodesic distances (shortest paths between points in the resulting graph), followed by a projection to a Euclidean space which preserves those distances.  As a result, ISOMAP uncovers the {\em intrinsic geometry} of the point cloud. Figure~\ref{fig:ISOMAPUnknotting} shows an example of running 2D ISOMAP on a trefoil knot, which is able to discover the single intrinsic loop in the complicated knot geometry.  In our example of the joint embedding of Doppler sensors, since we know an ideal embedding should lie on a line segment, we perform 1D ISOMAP.  As shown in column 3 of Figure~\ref{fig:DopplerSimExample1} and Figure~\ref{fig:DopplerSimExample2}, this leads to an SSM which more closely matches the speed profile than the joint embedding.  Also, the coordinate that is returned from 1D ISOMAP visually matches the speed profile better, up to a flip\footnote{Since ISOMAP seeks a Euclidean point cloud that respects geodesic distances only, the orientation and position of the returned point cloud are arbitrary.  So 1D ISOMAP in row 2 column 3 of Figure~\ref{fig:DopplerSimExample1} and Figure~\ref{fig:DopplerSimExample2} appears flipped.  But this is irrelevant, as the SSM is not affected by such a flip.}, which is factored out in the SSM.

\subsection{Normalization}
\label{sec:normalization}
There is an important issue of unit and scale that occurs when comparing any data across modalities.  In our case, we normalized the Doppler signals by their standard deviation before a joint embedding, but the scales of the resulting SSM are still most likely different from the scales of the speed SSM, especially due to the differences in dimension.  To cope with this, we apply two different normalization schemes to the SSMs before running IBDTW or persistence-based comparisons.  The first is to simply divide each SSM by its standard deviation, so that each matrix is left with a standard deviation of 1.  As shown in \cite{tralie2017warping}, however, a uniform scale is often not sufficient to address the changes that can occur between SSMs from different modalities.  Following \cite{tralie2017warping}, we also compare to a nonuniform histogram matching scheme to match an SSM A to an SSM B; that is, we discretize each SSM and construct a monotonic, one-to-one function $s$ so that when $s$ is applied to each pixel $A$, the histogram of $A$ approximately matches the histogram of $B$ (see \cite{gonzalez1992digital} Ch. 3.3).  We also come up with such a function from $B$ to $A$, and we keep whichever result yields a smaller score in our matching schemes.

\subsection{Isometry-blind Dynamic Time Warping (IBDTW)}
\label{sec:IBDTW}

\begin{figure}
\centering
\includegraphics[width=\columnwidth]{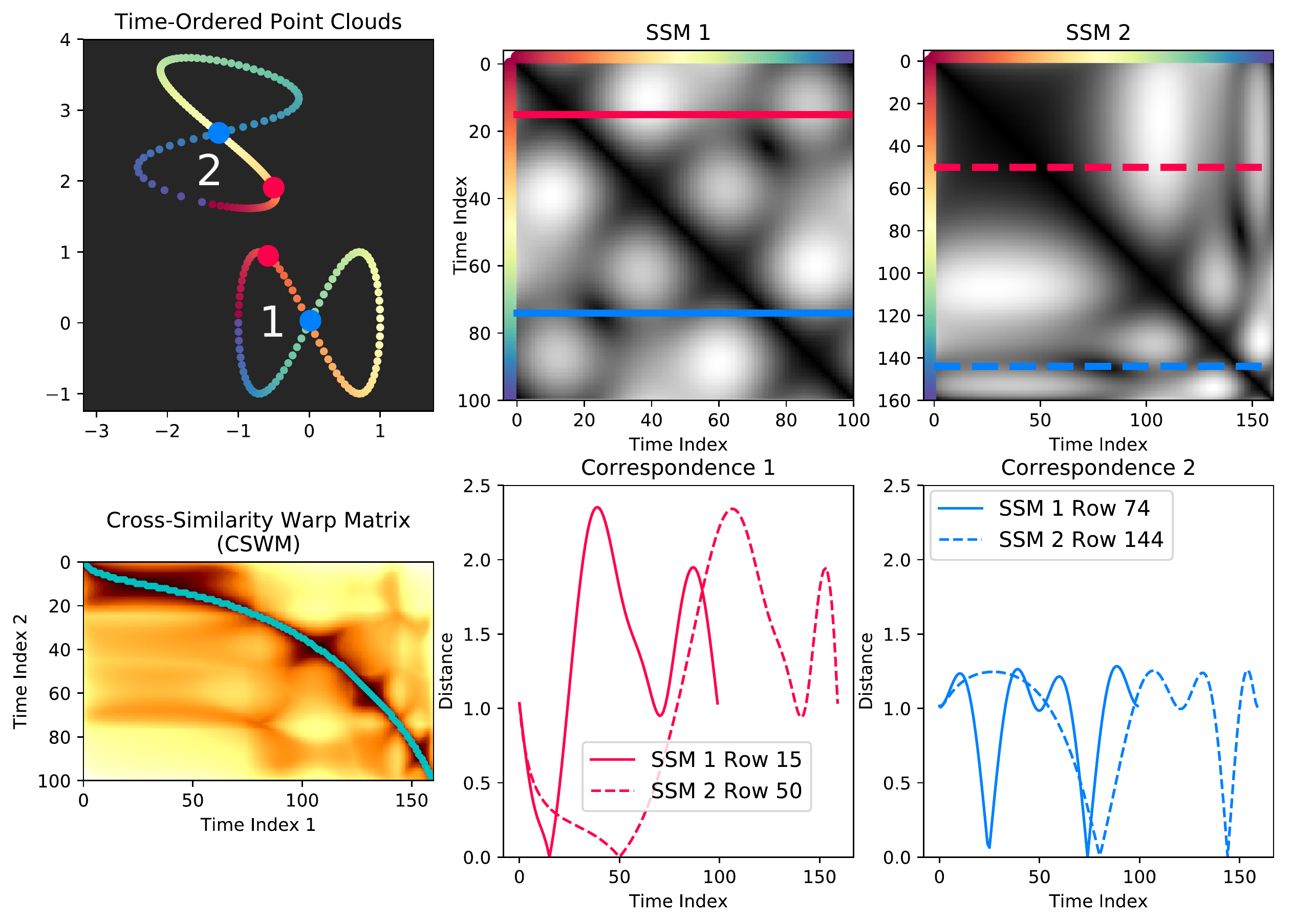}
\caption{A concept figure for isometry blind dynamic time warping (IBDTW).  The uncovered warping path between SSM 1 and SSM 2 is drawn in cyan in the lower left.}
\label{fig:IBDTWConcept}
\end{figure}

Once we have represented the speed profiles and Doppler as SSMs and normalized them appropriately, we need to match them to each other. Using SSMs to represent point clouds avoids a tricky spatial alignment. Nevertheless, the SSMs themselves sometimes still have to be aligned before they can be compared, due to differences in sampling of the point clouds.  Figure~\ref{fig:IBDTWConcept} shows an example of two point clouds sampled from a figure-eight curve which has been rotated/translated/flipped.  The spatial transformations do not affect the SSMs, but the second figure-eight point cloud has more samples than the first one, and they are also spread unevenly; more are towards the beginning (red points) than the end (magenta points).  As a result, SSM 2 is not a simple scale of SSM 1.  This can happen in the RF data if we want to match an ideal speed template SSM to the Doppler data from a driver who performs one of our actions, but who speeds up and slows down locally.  To address this, we apply a recently developed technique which is capable of uncovering local time parameterizations between SSMs  \cite{tralie2017warping}.  Figure~\ref{fig:IBDTWConcept} sketches the key observation underlying the approach.  For points which are in correspondence between the two point clouds, the corresponding rows of the SSMs are re-parameterizations of each other.  There is a well-known and proven algorithm known as dynamic time warping (DTW) for uncovering the parameterization between 1D time series (\cite{sakoe1970similarity,sakoe1978dynamic}, see also \cite{muller2007information} ch. 4), which can be applied between corresponding rows.  Since correspondences between rows are not known in advance, we try every row of the $M \times M$ SSM 1 against every row of the $N \times N$ SSM 2, and report the optimal cost of performing the DTW alignment in an $M \times N$ matrix known as the ``cross-similarity warp matrix'' (CSWM).  The optimal correspondence between the two SSMs can then be extracted as the shortest path from the upper left to the lower right of this matrix (lower left image Figure~\ref{fig:IBDTWConcept}, path shown in cyan), known as a ``warping path.''  The technique is referred to as ``isometry blind dynamic time warping'' (IBDTW).  For more details and theory, please refer to \cite{tralie2017warping}.  Since only a time warping needs to be uncovered, the technique is simpler and more effective than other multimodal alignment techniques which try to jointly solve for both a spatial transformation and a time warping \cite{zhou2009canonical,zhou2016generalized}.

\subsection{Persistence-based Comparison}
\label{sec:TDA}

In addition to the IBDTW, we also compare pairs of SSMs using techniques from topological data analysis (TDA).
First each SSM $S$ is summarized by the \emph{zero-dimensional persistence diagrams} $D_{+}(S)$ and $D_{-}(S)$ corresponding to its sublevel and superlevel set filtrations, respectively.
We briefly give an intuitive description of these diagrams here, beginning with an analogue for a lower-dimensional domain; see a TDA textbook (e.g., \cite{Edelsbrunner2010}) for a fully rigorous exposition, or one of many surveys (e.g., \cite{Motta2018}) for a discussion of TDA applications.
The idea of using persistence diagrams from super and sublevelset filtrations of time-ordered SSMs of isometric time-warped signals to compare them without alignment first appeared in the Ph.D. thesis of the first author in a more theoretical setting (\cite{tralie2017geometric}, Ch. 5.4), but this is the first known application of such an approach.

\begin{figure}
\centering
\includegraphics[width=\columnwidth]{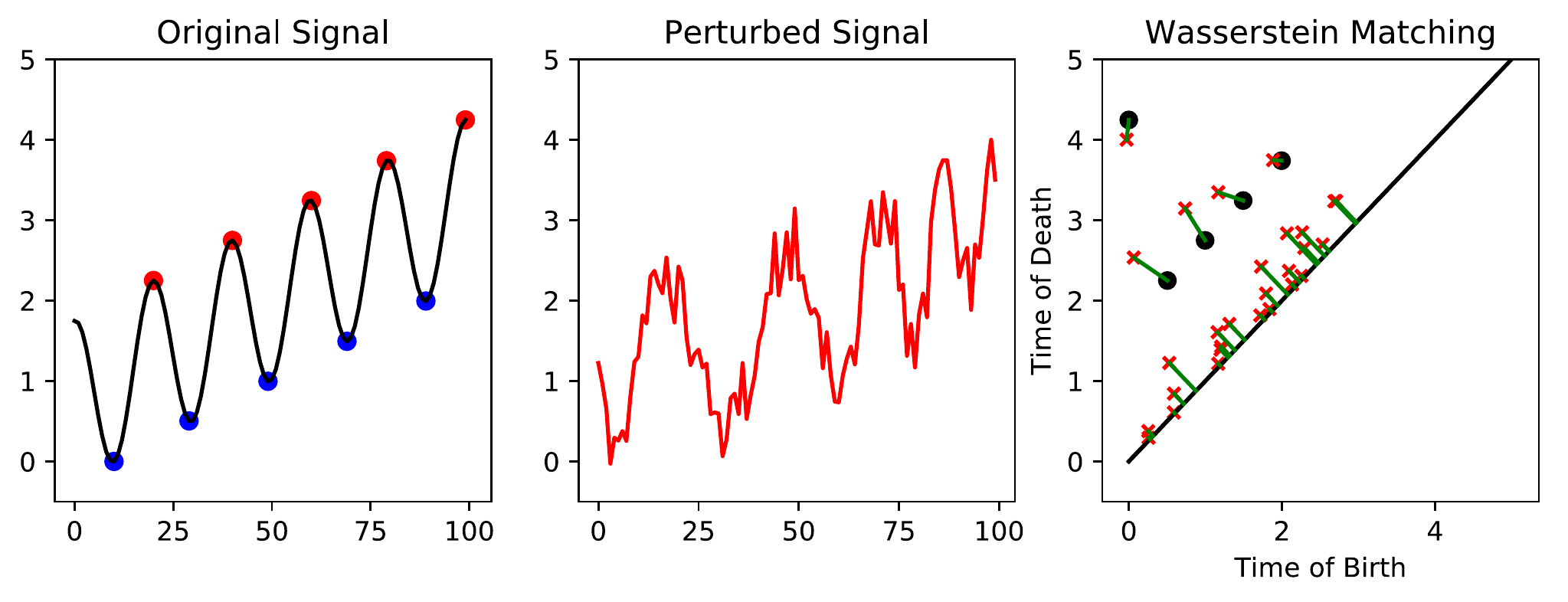}
\caption{An example of computing persistence diagrams of a sublevelset filtration of two 1D time series.  The second time series is both time warped and has added noise. The optimal Wasserstein matching between the two diagrams is indicated on the right.}
\label{fig:WassersteinConcept}
\end{figure}

Let $f:[a,b] \to R$ be a signal on some one-dimensional domain, such as the left panel of Figure \ref{fig:WassersteinConcept}.
The \emph{sublevel set} corresponding to a real-number threshold $\alpha$ is $\{x \in [a,b] \mid f(x) \leq \alpha\}$.
As these sublevel sets get bigger with increasing $\alpha$, starting with the empty set and ending with the entire domain $[a,b]$, connected components
appear (at local minima of $f$) and subsequently merge with other components (at local maxima of $f$).
The zero-dimensional persistence diagram $D_0(f)$, of $f$, summarizes this evolution as a multi-set of dots in the plane.
Each dot $u = (b,d) \in D_0(f)$ corresponds to a connected component which was created (born) at threshold value $\alpha = b$ and which then merged (died) at $\alpha = d$.
The diagram for our working example appears as the blue dots on the right side of the same figure.
Note that the persistence diagram of $f$ is entirely insensitive to any time-warping effects.

\begin{figure}
\centering
\includegraphics[width=\columnwidth]{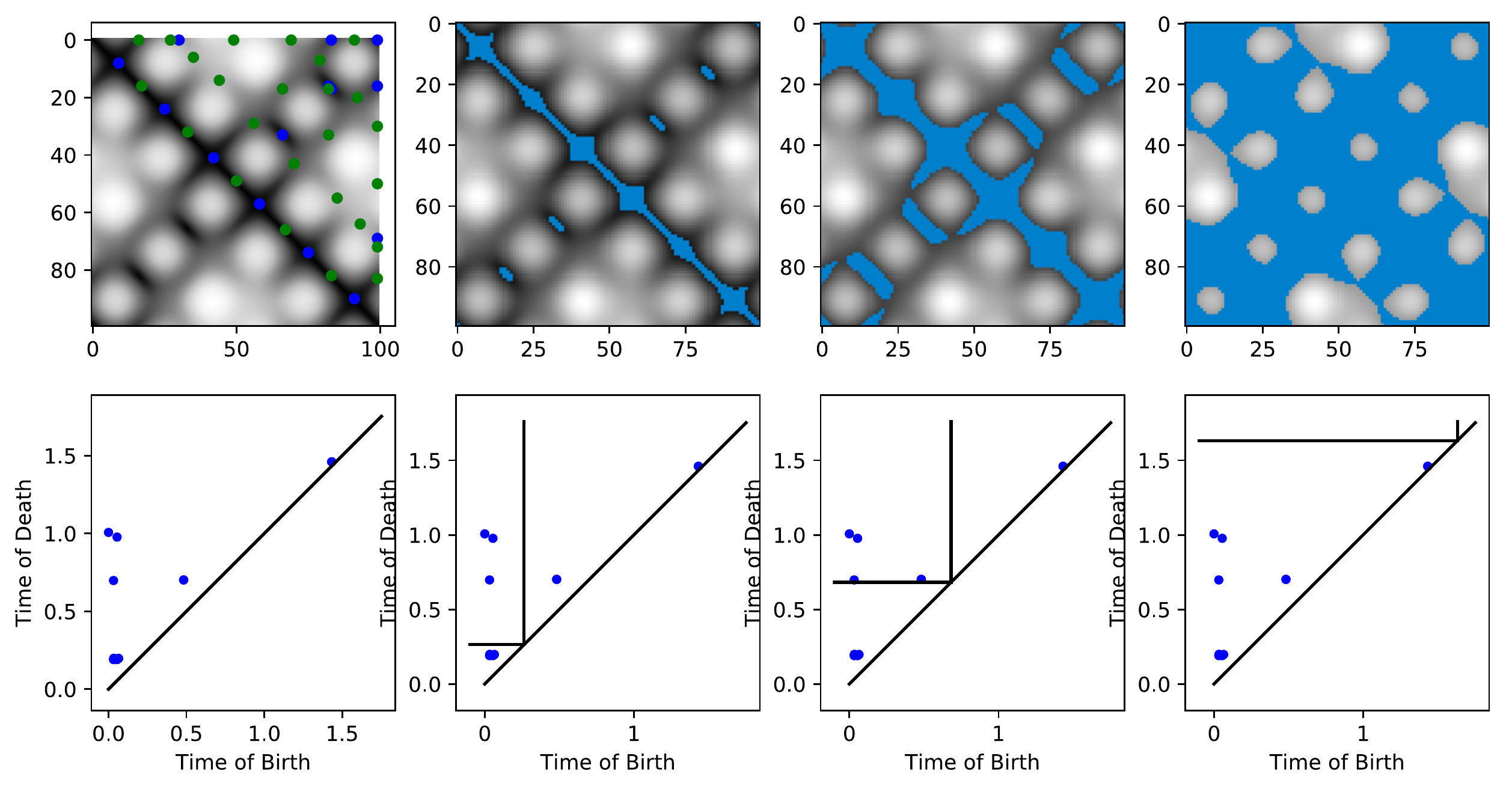}
\caption{An illustration of the watershed metaphor.  Local mins are drawn as blue dots on the upper left plot, and saddles where mins meet are are drawn as green points.  Different sublevel- set thresholds are shown in the plots on the top, and a line segment parallel to the y-axis and a line segment parallel to the x-axis meet at the diagonal.  The region in the upper left represents classes which are still alive.  In the middle plot, a few of the classes are about to merge together, and the younger ones die.  By the last level shown, all of the local mins have merge together, so there are no dots in the upper left region}
\label{fig:WatershedExample}
\end{figure}

The signal $g$ in the middle of Figure \ref{fig:WassersteinConcept} is a time-warped and noisy version of our $f$, and its persistence diagram appears as red crosses on the right.
These diagrams appear close, an observation formalized by the following typical metric used to compare persistence diagrams.
Given two persistence diagrams $P_1$ and $P_2$, the {\em $p$-Wasserstein Distance} $d^p(P_{S_1}, P_{S_2})$ is defined as

\[
d_W^p(P_1, P_2) = \inf_{\gamma \in \Gamma} \left( \sum_{i = 1}^{|\gamma|} ||P_1(i), \gamma(P_1(i))||^p \right)^{\frac{1}{p}}
\]

\begin{figure}
\centering
\includegraphics[width=\columnwidth]{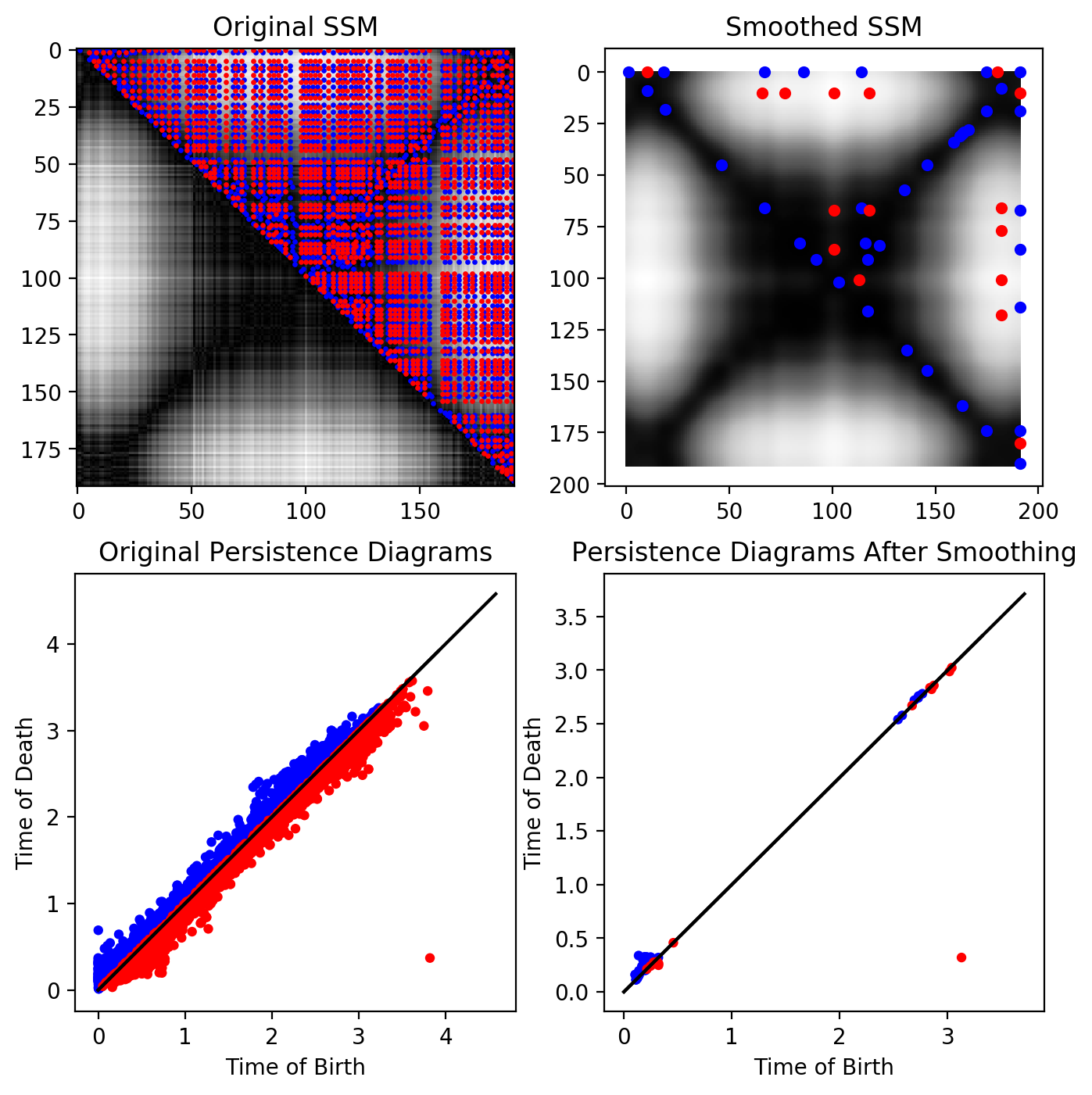}
\caption{An example of how a small amount of noise on the ``U Turn'' trajectory can lead to many low persistence points near the diagonal of the joint Doppler SSM.  To prevent the Wasserstein distance from blowing up, we perform a very slight smoothing on the image before computing the sub and superlevelset filtrations, which preserves the key features but reduces the noise near the diagonal.}
\label{fig:ImageSmoothing}
\end{figure}

where $\Gamma$ is the set of all perfect bipartite matchings (1 to 1 correspondences between points in $P_1(i)$ to those in $P_2(i)$), where all diagonal points are included in each with infinite multiplicity.  If $p = \infty$, this is known as the {\em bottleneck distance} between two persistence diagrams.  The bottleneck distance has been shown to be Lipschitz-continuous with respect to the Hausdorff distance between two point sets by \cite{CohenSteiner2007}, so the bottleneck distance is $L_{\infty}$ stable.  The same is not true of the $p$-Wasserstein distance in general, though the Wasserstein distance incorporates more information than simply the max distance, which can be more discerning in practice.  To address stability issues of the Wasserstein distance, we note that one common source of instability comes from noise blowing up points near the diagonal, all of which must be matched to the diagonal and add to the cost, as shown in Figure~\ref{fig:ImageSmoothing}.  To deal with this, we smooth each image by a Gaussian with standard deviation 3 pixels, which is enough to mitigate this effect, but not too much that important persistence information is destroyed (see \cite{chen2011diffusion}).

Of course, this paper is concerned with comparing SSMs, which are two-dimensional images, not one-dimensional signals.
Nonetheless, the definitions above extend to the higher-dimensional case, where we think of an SSM as a real-valued function $f$ on a square domain (since our SSMs are symmetric,
we can actually just imagine the domain to be the upper triangle).
This is perhaps best understood by the \emph{watershed} metaphor (Figure \ref{fig:WatershedExample}), where we imagine a giant flood on a terrain and $f$ indicates the water height at a given domain point.
In this case, new components in the sublevel set appear at local minima of $f$, which we imagine to be the bottom of valleys, and these components merge at saddle points of $f$, thought of as mountain passes. 
Just as in the one-dimensional signal case, this component evolution is summarized in a persistence diagram $D_{+}(f)$.
The same process can be repeated for the negative of the function, resulting in the superlevel set persistence diagram $D_{-}(f)$.
In this case, we compare two SSMs by taking the sum of the Wasserstein distances between their sublevel set diagrams and between their superlevel set diagrams.

\begin{figure*}
\includegraphics[width=\textwidth]{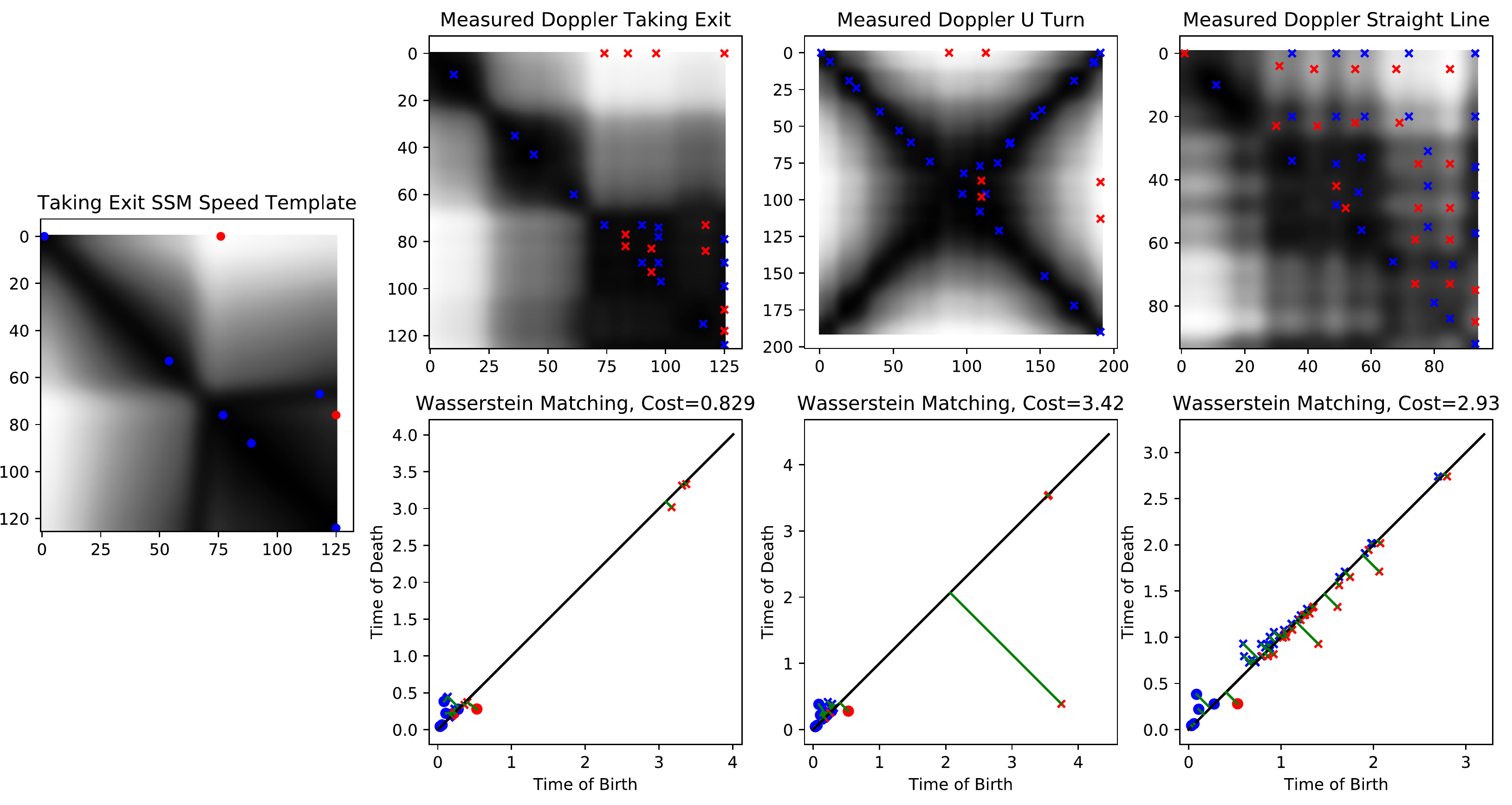}
\caption{Examples of matching the Z-normalized SSM of a U Turn speed template to the Z-normalized, smoothed SSMs from various individual Doppler sensors.  Local mins are drawn in blue, and local maxes are drawn in red.  Critical points in the speed SSM are drawn as dots, and critical points in the Doppler SSMs are drawn as Xs.  Only critical points on the upper triangular part of each matrix are considered, as the matrices are symmetric.  For the true class type of taking an exit (left column), there is a bit of measurement noise that leads to some small persistence dots, but the overall Wasserstein match is of low cost.  For an incorrect action of making a U turn (center column), the action is missing a large local max, so the cost is higher.  For the incorrect action of going straight (right column), the velocity hardly changes, so the noise has a large effect, leading to many critical points; this ultimately leaves a higher score in the optimal match, as most of these critical points must match to the diagonal.}
\label{fig:WassersteinMatchingDoppler}
\end{figure*}

\section{Experiments and Results}
\label{sec:ExpRes}


We now examine how well our different SSMs from simulated 2GHz\footnote{Note that in our simplistic simulation where we neglect decaying amplitude and other factors, the choice of carrier frequency does not impact the results, as we normalize the Doppler shifts.  But we chose a 2Ghz frequency for the sake of consistency with real world scenarios.} Doppler sensors yield true positive and true negative comparisons with speed profiles.  To assess this, we compute speed SSMs and Doppler SSMs between 3 actions: going straight, taking an exit, and making a U-turn, as shown in Figure~\ref{fig:Trajectories}.  The radio signals are received by four detectors at the corners of a square of side length 4 kilometers.  We use SSMs from perfect speed traces as queries to Doppler SSMs.  We generate 100 different Doppler measurement runs for each action.  We randomly rotate/translate/flip the trajectories for each draw, and we add Gaussian noise to each receiver with a standard deviation equal to $3\%$ of the maximum Doppler shift, yielding an average pSNR of 51.6dB.  An example is shown in Figure~\ref{fig:NoiseExample}.

After running the 100 draws for each of the 3 actions, we generate SSMs for comparison.  Since there are multiple receivers, we compared four different ways of incorporating them:
\begin{itemize}
\item Using the SSM from each receiver individually (Ind)
\item A simple average of the SSMs from each receiver (Ind Avg)
\item A joint Euclidean embedding of all of the sensors (Joint)
\item ISOMAP of the joint embedding of all of the sensors (ISOMAP)
\end{itemize}

We also explore deviation based normalization (Std) and histogram matching normalization (HistMatch), as described in Section~\ref{sec:normalization}.

\begin{figure}
\centering
\includegraphics[width=\columnwidth]{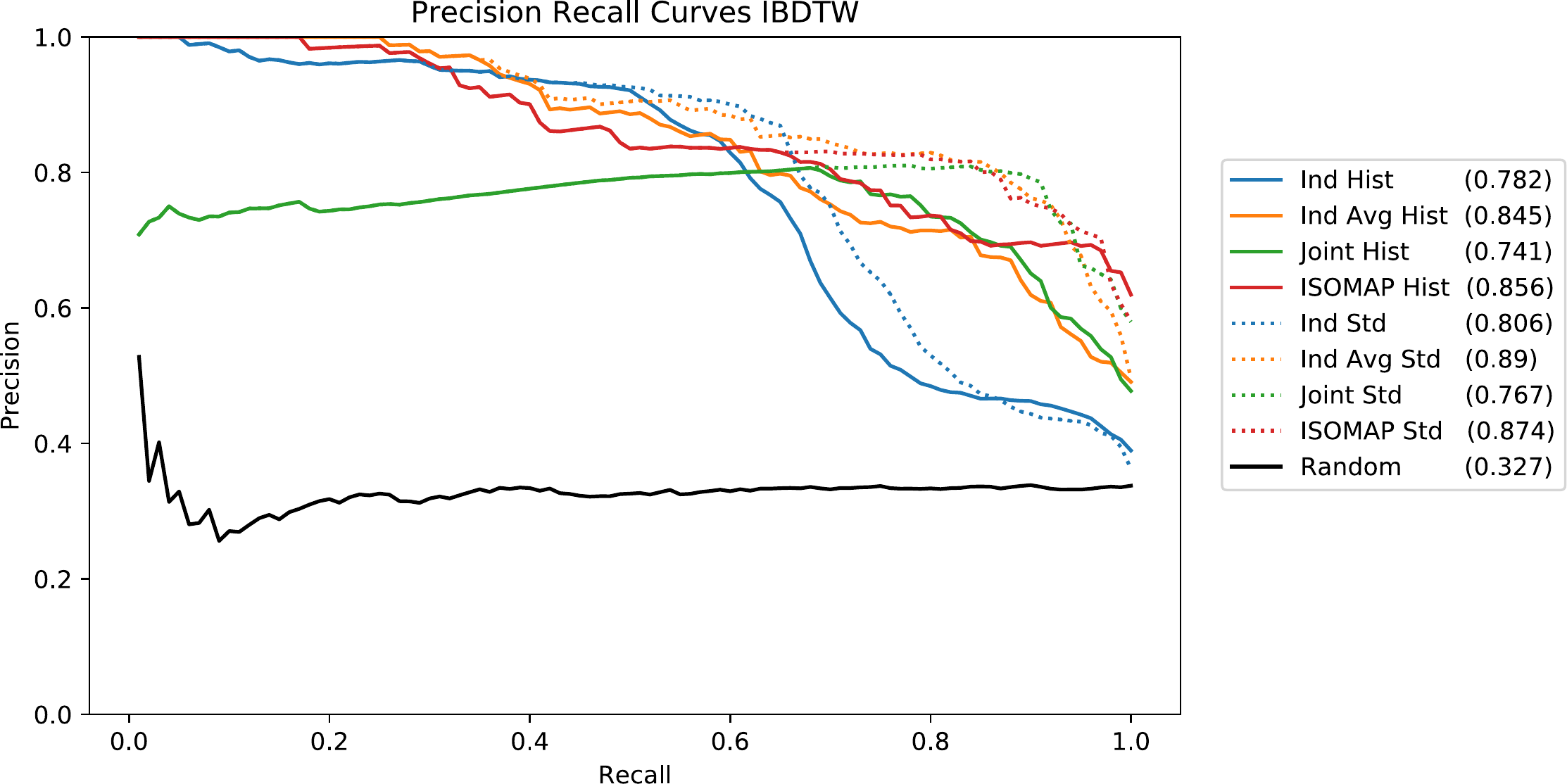}
\caption{Precision recall curves for different techniques using the IBDTW alignment technique.  The mean average precision (MAP) for each technique is shown in parentheses in the legend.}
\label{fig:PRIBDTWs}
\end{figure}

\begin{figure}
\centering
\includegraphics[width=\columnwidth]{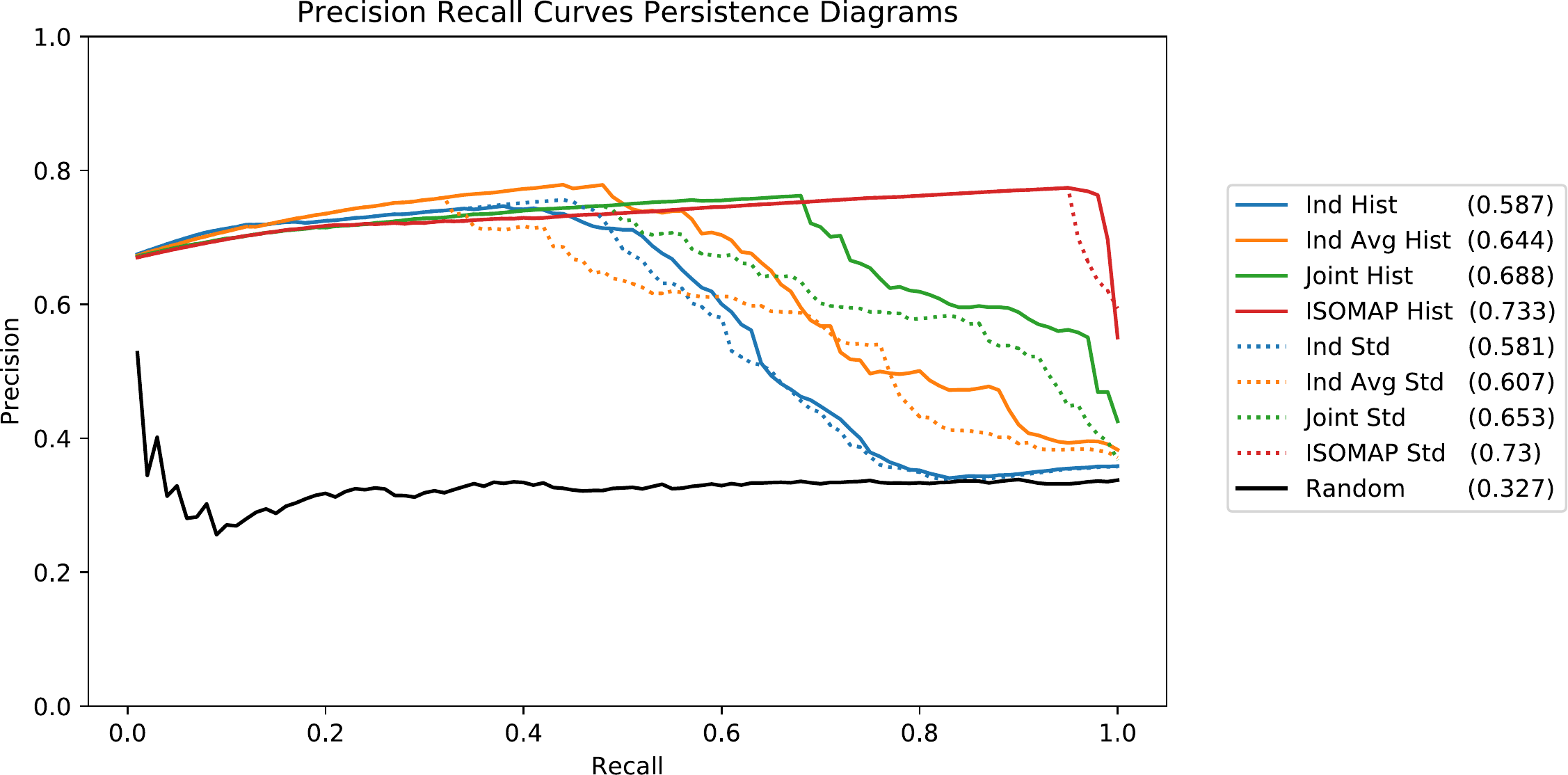}
\caption{Precision recall curves for different techniques using the Wasserstein distance between persistence diagrams.}
\label{fig:PRPDs}
\end{figure}

For each of the eight combinations of SSM computation and normalization, we measure a similarity between the three ideal speed SSMs and all of the sampled Doppler data from every trajectory, for a total of 900 comparisons.  For each motion, we compute precision recall curves after sorting the Doppler SSMs from most to least similar with respect to the speed SSMs, using IBDTW and Wasserstein distance between.  We smooth the images before computing persistence diagrams to control for noise blowing up near the diagonal, as discussed in Section~\ref{sec:TDA}, but we do not smooth them before IBDTW.  Figure~\ref{fig:PRIBDTWs} shows the precision recall curves for all of these experiments with all of the methods using IBDTW, and Figure~\ref{fig:PRPDs} shows the precision recall curves using the Wasserstein distance between persistence diagrams.  We note that all techniques are better than random guessing (which has a mean average precision of about 1/3), but fusing information from all of the sensors always out-performs the individual sensors.  Furthermore, ISOMAP on a joint embedding always improves results over the joint embedding alone.

In our experiments, we also found that adding a small amount of noise yields better results than noise at all.  Figure~\ref{fig:ImageSmoothing} suggests why this might be.  For the car moving in a straight line, there is little variation in the speed, so any noise added has a much larger variance than variation in the speed.  After normalization, this leads to large mins and maxes in the SSM.  By contrast, the ratio of the motion variation to the noise variance in the other two motions, where speed is changing, is much larger, so the noise has a lesser effect.  This highlights one of the potential issues of trying to identify a scenario where ``nothing is happening.''


\section{Multiple Hypothesis Tracking}
\label{sec:MHT}

Many systems employ sensors to interpret the environment, typically one involving one or more moving targets, such as vehicles.
The target-tracking task is to gather sensor data from the environment and then to partition these data into tracks that are produced by the same target.
A key challenge is to ``connect-the-dots'': more precisely, to take a sensor observation at a given time and associate it with a previously-existing track (or to declare that this is a new object). This is especially challenging when, as often happens in cluttered urban environments \cite{Demars2015}, a needed sensing modality ``goes blind'' for a period of time.
We describe here, in very broad strokes, a potential way to integrate the ideas presented in this paper into multi-target tracking methodology.
The key idea is that a large suite of experiments such as the ones in Section \ref{sec:ExpRes} can build a \emph{dictionary} between simple motion primitives and their SSMs as measured
by a disparate array of sensing modalities.

There are a number of high-level designs for multi-target multi-sensor tracking systems, but most contemporary ones fit within the multiple hypothesis (MHT) paradigm
(\cite{Reid1979} \cite{Blackman1999}).
Typical MHTs formulate the 'connect-the-dots' problem as one of Bayesian inference, with competing multi-track hypothesis receiving scores; e.g., with the Bayesian Log-Likelihood
Ratio of Bar Shalom et. al. \cite{Barshalom2007}.
A deferred decision logic scheme is used to process observed data in recursive fashion: parent hypotheses are updated with new data to form child hypotheses, which are then subsequently pruned, using thresholds on hypothesis score and/or total number of hypotheses, to provide a reasonable number of parent hypotheses for the next round of observations.

The track file of pruned candidate tracklets can become much larger than the actual number of targets, where the tracklets are alternative attempts to represent targets with the same data.
Two of the authors were involved in a recent OSD-funded project \cite{OASIS} which showed that periodic appraisals of behavior, using a description of shape simpler than but similar to SSMs, applied to motion primitives of tracklets, were able to dramatically reduce the number of candidate tracklets and thus massively improve tracker performance in simulated tests(\cite{Bendich2016}, \cite{Rouse2015SPIE}).
This work analyzed motion primitives directly from imagery (in this case, WAMI) modalities.
But the SSM-based ideas developed here will potentially allow for the same analysis from entirely different modalities.

The following notional scenario illustrates the potential benefits of this proposed effort. Suppose that a standard MHT is maintaining a track of one vehicle that is going along a straight path, as in the third row of Figure \ref{fig:Trajectories}. Suddenly, the imagery sensor goes blind, or simply focuses its attention elsewhere, for a sequence of consecutive frames
The MHT can do nothing other than continue forward the purported vehicle track, with an ever-expanding uncertainty ellipse.
During this time, the vehicle exits the highway, as in the first row of Figure \ref{fig:Trajectories}.
When vision returns, the MHT attempts to find the vehicle, but it is ignorant of the exit. It will probably conclude that the original track has terminated; if it observes the original car along its new trajectory, it will probably just start another track. In either case, the data-to-track association problem will not be correctly solved.
On the other hand, suppose that the MHT is able to make use of RF data the entire time. During the period of blindness, a quick SSM calculation, plus a dictionary lookup of a coarse resized SSM disregarding local time warping (see the top row of Figures \ref{fig:DopplerSimExample1} and \ref{fig:DopplerSimExample2}), will present the tracker with at least a reasonable series of options, one of which will be ``the vehicle has probably exited the highway.''
Put more rigorously, classification results analogous to those presented in Section \ref{sec:ExpRes} will lead to relative likelihoods for a set of potential primitive motions.
At the very least, the MHT will be able to create a new hypothesis that, when coupled with the imagery observations after vision returns, will allow for the correct continuation of the track.

\section{Conclusions}

This paper introduced the idea of using self-similarity matrices (SSMs) for comparison of aerial data streams arising from distinct sensing modalities.
In addition to theoretical development and exposition, we show via proof-of-principle experiments that our techniques have promise for passing information between modalities about specific primitive motions within vehicle trajectories.

Although the simulations we undertook are under a simplified set of assumptions, many complexities still arise, and we show the benefits of a geometric way of thinking for tackling this complexity.  Now that we have laid the conceptual groundwork, we would like to apply the techniques to more complicated, real world data.
We will also expand these methods to comparisons between other sensing modalities, such as infrared and acoustic sensors.
Finally, we also plan to implement the ideas laid out in Section \ref{sec:MHT}, in order to demonstrate the benefits of this geometric approach for cross-modal vehicle tracking.

\newpage

\bibliographystyle{plain}
\bibliography{GDAreferences}

\end{document}